\documentclass[11pt,a4paper]{article}
\usepackage[hyperref]{emnlp2020}

\usepackage{hyperref}

\usepackage{times}
\usepackage{latexsym}
\usepackage{url}
\usepackage{dirtytalk}

\usepackage{algorithm}
\usepackage{algorithmicx}

\usepackage{amsfonts}
\usepackage{amsmath}
\usepackage{amssymb}
\usepackage{graphicx}
\usepackage{multirow}
\usepackage{booktabs}
\usepackage[shortlabels]{enumitem}

\usepackage{array}
\newcolumntype{H}{>{\setbox0=\hbox\bgroup}c<{\egroup}@{}}

\aclfinalcopy 

\newcommand{\dsname}{NYTWIT}

\title{Will it Unblend?}

\author{Yuval Pinter \\
  School of Interactive Computing \\
  Georgia Institute of Technology \\
  Atlanta, GA, USA \\
  \texttt{uvp@gatech.edu} \\\And
  Cassandra L. Jacobs \\
  Department of Psychology \\
  University of Wisconsin \\
  Madison, WI, USA \\
  \texttt{cjacobs2@wisc.edu} \\\And
  Jacob Eisenstein \\
  Google Research \\
  Seattle, WA, USA\\
  \texttt{jeisenstein@google.com} \\}

\date{}

\begin{document}
\maketitle
\begin{abstract}
    Natural language processing systems often struggle with out-of-vocabulary (OOV) terms, which do not appear in training data.
Blends, such as \textit{innoventor}, are one particularly challenging class of OOV, as they are formed by fusing together two or more bases that relate to the intended meaning in unpredictable manners and degrees.
In this work, we run experiments on a novel dataset of English OOV blends to quantify the difficulty of interpreting the meanings of blends by large-scale contextual language models such as BERT.
We first show that BERT's processing of these blends does not fully access the component meanings, leaving their contextual representations semantically impoverished.
We find this is mostly due to the loss of characters resulting from blend formation.
Then, we assess how easily different models can recognize the structure and recover the origin of blends, and find that context-aware embedding systems outperform character-level and context-free embeddings, although their results are still far from satisfactory.

\end{abstract}

\section{Introduction}


For the token-based architectures that dominate contemporary natural language processing, a particularly difficult form of linguistic generalization arises from unseen phenomena at the word level, where novel sequences of characters, morphemes, or phonemes are known as out-of-vocabulary (\textbf{OOV}) terms~\cite{brants2000tnt,plank2016non,heigold2017robust}.
Pretrained transformers like BERT~\cite{devlin-etal-2019-bert} handle OOV terms by \textbf{subtokenization}: segmenting all whitespace-delimited tokens into smaller units, from which any OOV term can be constructed~\cite{sennrich-etal-2016-neural}.\footnote{Another approach is to operate directly at the character level~\citep[e.g.,][]{ling2015finding}, but this has not been widely adopted within the mainstream framework of transformer-based models~\cite{vaswani2017attention}, due in part to the difficulty of scaling transformer-based models to sufficiently large contexts when operating at the character level.}
But while this approach is well suited for phenomena like concatenative English morphology, many linguistic processes generate OOV terms that cannot be cleanly decomposed into meaningful subtoken segments.

In this paper we address a particularly interesting and challenging source of OOV terms: novel \textbf{blends}~\cite{algeo1977blends}, also known as portmanteaux~\cite{deri2015make}.
Blends are constructed from the combination of multiple \textbf{bases} into a new form, in which some characters is shared across both bases: for example, \emph{shop} + \emph{optics} = \emph{shoptics}. 
In this way, blends differ from other lexical compounds (e.g., \emph{watermelon} = \emph{water} + \emph{melon}), which are formed by simple concatenation.
Examples of OOV blends and their bases from our novel English blends dataset, collected from a natural source linked to the blends' originating contexts (\S\ref{sec:data}), are presented in \autoref{tab:sample_blends}.
OOV blends are especially challenging to process, due to their combination of function-level semantic novelty with the form-level pathology of an unexpected character sequence.
\begin{table*}
    \small
    \centering
    \begin{tabular}{lllHll}
    \toprule
        Blend & PAXOBS & Bases & Linearity & Semantic relation & Definition \\
        \midrule
        \underline{hatriotism} & {\tt AXXBBBBSSS} & hate patriotism & \textsc{linear} & \textsc{attribute}  & Hate disguised as patriotism. \\
        \underline{shoptics} & {\tt AAXXBBBS} & shop optics & \textsc{linear} & \textsc{loc-part-whole} & The social image projected when shopping. \\
        {innoventor} & {\tt XXAAXBBXXX} & innovator inventor & \textsc{nonlinear} & \textsc{causal} & A person who innovates by inventing. \\
        {thrupple} & {\tt AAABOBBB} & three couple & \textsc{nonlinear} & \textsc{containment} & A group of three people acting as a couple. \\
         \bottomrule
    \end{tabular}
    \caption{A sample of the blends from the dataset, with definitions and our full annotation as described in~\S\ref{sec:data}.
    Linear blends are underlined.
    }
    \label{tab:sample_blends}
\end{table*}

Our main contribution is to offer what is to our knowledge the first analysis of how transformer-based contextual embedding models process novel blends and the representations they are able to produce for these challenging forms.
First, we examine the impact of blends' wordforms by comparing the ability of contextualized models to represent blends against the minimally-different case of novel lexical compounds.
In~\S\ref{sec:motiv}, we show that the limited ability of contextual language models to represent novel blends' components faithfully is primarily attributable to their form properties, whereas semantic differences between compounds and blends play a much smaller role.
We then investigate how well several methods are able to recover the morphological boundaries within blends, which could mitigate the impact by splitting blends into segments contributed by each constituent base (\S\ref{ssec:segmentation}).
Finally, we attempt to recover the constituent bases given a segmentation (\S\ref{ssec:recovery}). 
Even under favorable conditions, we find that systems proposed previously for similar tasks struggle on blends, showing limitations of form-based and distributional similarity approaches.
We propose a novel unsupervised base recovery method using contextualized masked language models, \textsc{BERT ranker}.
While this system performs well relative to others, we find substantial room for improvement.
In our view, these results demonstrate the need for future work on our novel dataset and associated tasks.\footnote{We release our code and data at \url{http://github.com/yuvalpinter/unblend}.}

\section{Complex Words Dataset}
\label{sec:data}

Our proposed investigation of the behavior of NLP systems on novel complex words requires a high-quality, reliable resource of truly novel blends and compounds in their original contexts, annotated for character sequence composition and semantic properties.
The \dsname{} dataset~\cite{pinter2020nytwit} contains English words new to the New York Times extracted by a bot\footnote{\url{www.twitter.com/NYT\_First\_Said}} between the dates of November 2017 and March 2019 with associated news article contexts.
Words were annotated for their type of novelty.

We extract and further annotate three types from this dataset (version 1.1): blends (142 items), transparent compounds (121), and opaque compounds (49).\footnote{Originally annotated as \say{compositional} and \say{new} compounds, respectively.}
The difference between the compound classes is semantic and somewhat subjective: transparent compounds have meanings which are comprehensible with little context (e.g.~\emph{quizmaker}, a person who makes quizzes), while opaque compounds exhibit metaphoric or allusive semantics (e.g.~\emph{deathbox}, a dangerous car).

The first two authors annotated each word for its constituent bases, the character locations in which each base is represented, and the semantic relation between the bases.
A sample of annotated blends is presented in~\autoref{tab:sample_blends}.
All disagreements resulting from the first round of blend base annotation (7\%) were resolved by discussion, with the help of the words' originating context.
These contexts vary considerably in their length and informativity,\footnote{Compare, for example, the following context sentences: \say{\emph{Blaspy}?}; \say{The \emph{procrastibaker} must believe that it is possible to be simultaneously working on a document, buttering pans and separating eggs.}} but typically contain direct or indirect disambiguating information, and sometimes the component bases themselves: for blends, 40.3\% of documents contained at least one of the bases within sentences where the blends appear (e.g. only \emph{shop} or \emph{optics}, for \emph{shoptics}), while 10.2\% contained both.

\paragraph{Semantic relations.}
An author annotated all blends and compounds in the dataset according to the well-studied semantic taxonomy of \newcite{tratz-hovy-2010-taxonomy}, which was designed for multiword nominal compounds (e.g.~\emph{cooking pot}).
These relations were not intended to be applied to other types of phrases, blends, or lexical compounds.
However, by referring to the official taxonomy, the expanded definitions in \newcite{dima-hinrichs-2015-automatic}, and the coarse- and fine-grained relation training data, the annotator was able to assign one of twelve coarse-grained relation classes to each word.\footnote{For example, \textsc{attribute} was applied to the adjective-noun blend \emph{fitfluencers}.
69\% of the words contain a non-noun base.}

As a preliminary check,
we trained a relation classifier following the approach of \newcite{dima-hinrichs-2015-automatic}, a single-hidden-layer classifier over GloVe embeddings~\cite{pennington-etal-2014-glove}, on the \textsc{random} partition of the \newcite{tratz-hovy-2010-taxonomy} data.\footnote{The model trained on this split, set to $d_e = d_h = 50$, slightly outperformed an identical one trained on the \textsc{lexical} split, and its test set accuracy on the original dataset is $.721$, close to replication. The \textsc{lexical} split was created to correct an over-representation of some compound bases in \textsc{random}, one which biases statistical models toward lexical memorization (see details in \S 4.1 of \newcite{shwartz-waterson-2018-olive}), but has no bearing on our dataset.}
This model achieved $.203$ accuracy and $.173$ macro-F1 on our dataset for all 311 items, substantially higher than baselines such as majority class ($.087$ acc. / $.013$ F1) and random prediction calibrated to the marginal label distributions ($.106$ acc. / $.078$ F1 for the best of ten runs), indicating credible annotation.\footnote{The numbers for blends only are $.148$ / $.079$ vs. $.063$ / $.020$ for majority class.}
This performance is still poor relative to multiword compounds, possibly due to the fundamentally different linguistic processes governing lexical compounding and blending processes as opposed to multiword compounding.

\paragraph{Character-level labels.}
We introduce a character-level labeling schema to help classify blend types and evaluate and train blend segmentation models, called \textbf{PAXOBS}.\footnote{May be pronounced like \say{pack sobs}.} 
Each character is labeled as \texttt{P} or \texttt{S} if it is in a prefix or suffix, respectively; as \texttt{X} or \texttt{O} if it is contributed by more than one or none of the bases, respectively; and by successive letters of the alphabet for characters from only each base, typically just \texttt{A} or \texttt{B}.\footnote{This framework is loosely similar to \emph{edit scripts}~\cite{chrupala-etal-2008-learning}, but rather than transducing one string into another, the task is to combine two strings into a third.}
This schema covers the full range of processes undergone in blending, except for annotation of characters removed altogether from the bases (e.g.~the \textit{e} from \textit{hate} in \emph{hatriotism}),
and may be trivially applied to lexical compounds as well.

Blends may be classified into further subcategories based on the correspondence between their form and the bases.
For example, \textbf{linear} blends are similar to compounds, as each base's portion appears uninterrupted in the blend (underlined in \autoref{tab:sample_blends}).
Formally, a blend is linear if its label sequence contains: no \texttt{O}; no \texttt{A} preceded by a \texttt{B} or \texttt{X}; and no \texttt{B} followed by an \texttt{A} or \texttt{X}.
In our dataset 59\% of blends are linear, though prior work has reported up to 95\% linear blends among blends extracted from a curated lexicon~\cite{cook-stevenson-2010-automatically}.
One possible explanation for this discrepancy is that words that make it into common use may be simpler in their surface quality.

\section{Blends in Context}
\label{sec:motiv}

\begin{figure*}
    \centering
    \small
    \begin{tabular}{cc}
        \includegraphics[width=0.48\linewidth]{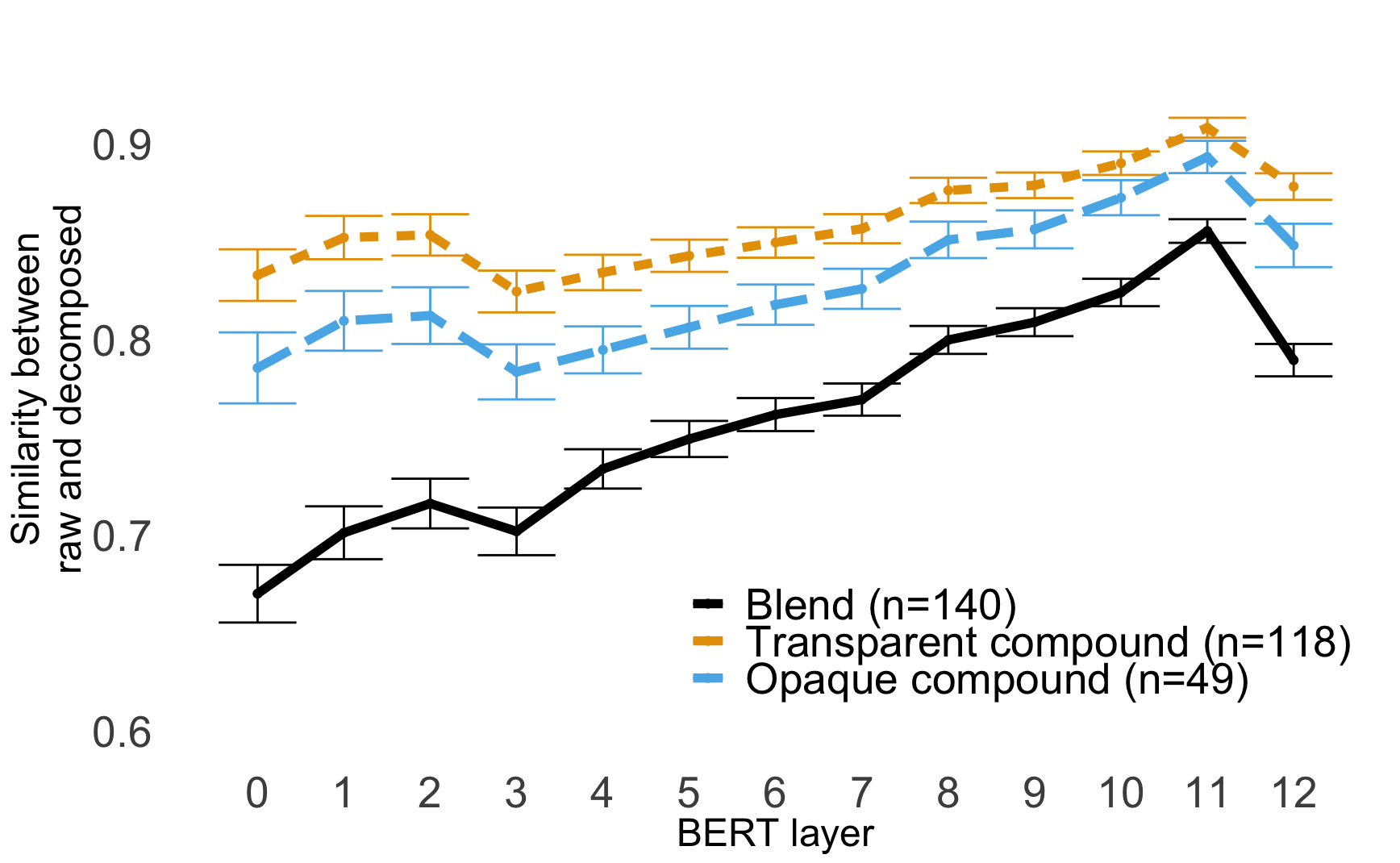} & 
        \includegraphics[width=0.48\linewidth]{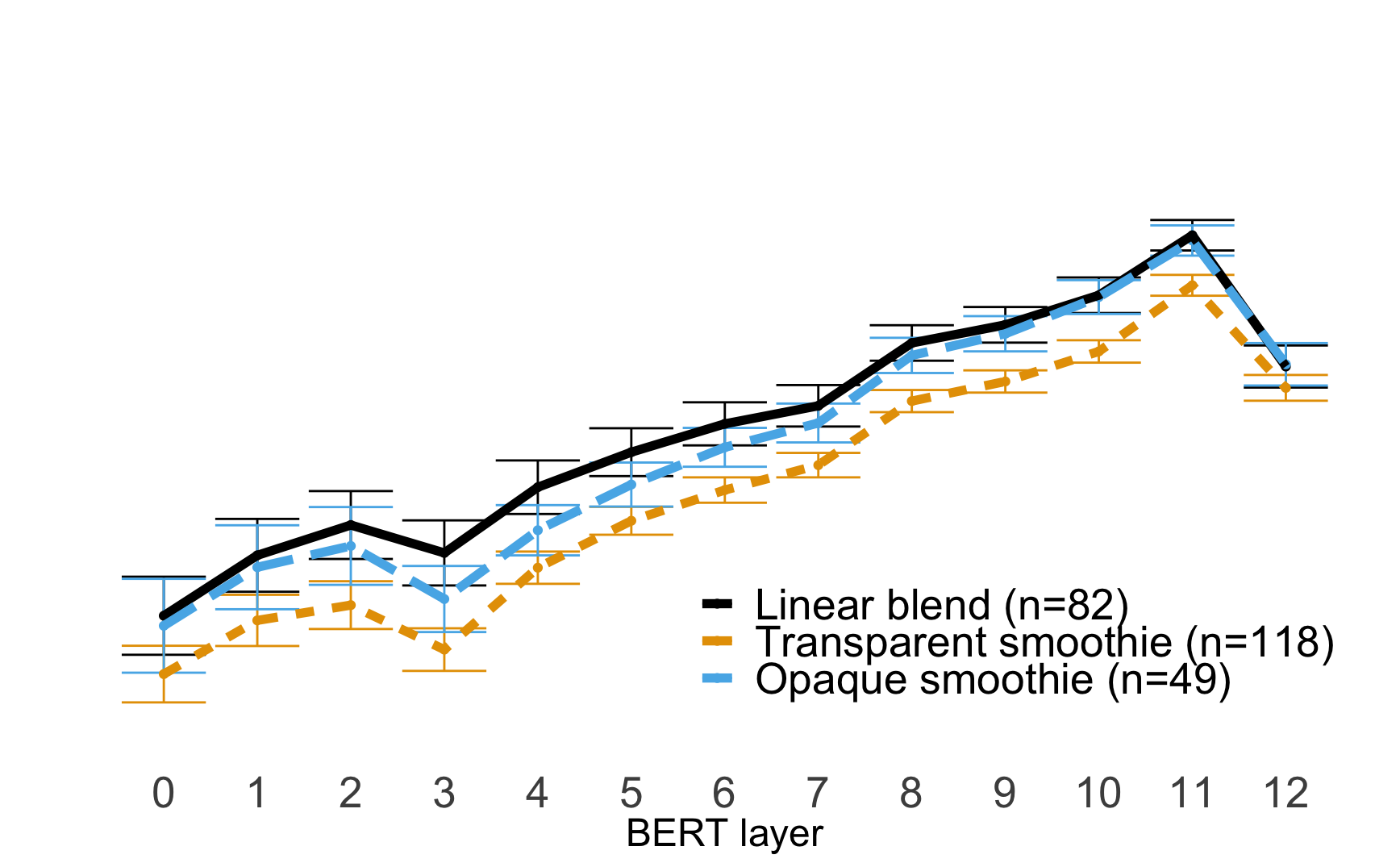} \\
        (a) & (b)
    \end{tabular}
    \caption{Pretrained BERT's layer-wise similarity between representations of (a) complex OOVs and their base components; and 
    (b) linear blends and \say{smoothies} (\S \ref{ssec:comp-smooth}), lexical compounds forced to lose characters while remaining linear.
    All representations are computed using the original context in which the words appear.
    Error bars represent standard error of the mean over the class.}
    \label{fig:bert_sims}
\end{figure*}

Novel blends are a unique linguistic phenomenon, posing challenges for automated systems on many different levels.
However, the sparsity of their appearances in real-world text, as well as the expertise required for creating a natural language understanding task which uses specific documents from a large variety of domains as supporting information, make the evaluation of the effect of novel blends on this type of downstream task an impractical goal for the scope of this work.
Instead, we assess the treatment of novel blends at the representational step of contemporary contextualized language models, by performing an analysis of their processing by BERT~\cite{devlin-etal-2019-bert}.
To gauge how well BERT represents blends, we conduct a comparison with its treatment of a minimally-different control class of novel words, namely \textbf{lexical compounds}.
These are forms where at least two bases are concatenated in full (e.g.~\emph{\textcolor{brown}{quiz}\textcolor{blue}{maker}}), without the character loss incurred in blends.

Our analysis begins with the assumption that in any given context, the meaning representation of a complex word (blend or lexical compound) must be \textbf{composed} from its bases, which we can estimate using the representational similarity between a complex word and its bases in the same context.
This criterion can be viewed as a form of linguistic generalization, and if satisfied, enables downstream models to produce consistent results across related words.
To test this criterion, we compute the vector similarities between the contextualized representations of complex words and their components, a method that coheres with human judgments of contextual semantic similarity~\cite{giulianelli2020analysing}.
We probe BERT\footnote{We use the \texttt{base-uncased} flavor and the Huggingface implementation~\cite{wolf2019transformers} throughout the paper.} with synthetic inputs constructed by replacing each complex word with its space-delimited bases.
Formally, given a sentence $S=(w_1, \dots, w_{i-1}, x, w_{i+1}, \dots, w_n)$ where $x$ denotes a blend or compound with contributing bases $b_1, b_2$, we record the average vector across $x$'s wordpiece tokens for each layer output in BERT's transformer stack, $e^{(l)}(x), l\in 0, \dots, 12$, 
and compute its cosine similarity with the averaged vectors $\frac{1}{2}(e^{(l)}(b_1) + e^{(l)}(b_2))$ in the sequence $S'=(w_1, \dots, w_{i-1}, b_1, b_2, w_{i+1}, \dots, w_n)$.\footnote{Four forms in our dataset have three bases: \say{fanimatic} = \textit{fan} + \textit{animation} + \textit{cinematic}, \say{shaggydoodle} = \textit{shaggy} + \textit{labrador} + \textit{poodle}, \say{frenemesis} = \textit{friend} + \textit{enemy} + \textit{nemesis}, \say{orchaestraits} = \textit{orca} + \textit{orchestrates} + \textit{straits}. In these cases, we include the vectors for all three bases.
One blend, \say{pregret}, has only one base, against which it is compared.

Five words are missing from the analysis as they no longer appeared in their original contexts at scraping time due to editorial actions on the NYT website: the blends \say{humailiation} and \say{crapberg}; and the compounds \say{cybersensation}, \say{garagerock}, and \say{storytale}.}

\autoref{fig:bert_sims}(a) compares the per-layer similarities for blends with the two types of compounds described in~\S\ref{sec:data}:
We find a clear distinction between blends and both compound classes.
For compounds, BERT induces representations that are very similar to those of the components at all layers of the model.
For blends, these representations diverge greatly, especially in the lower layers of the model, which capture surface-form characteristics of the input~\cite{jawahar-etal-2019-bert}.

Since the difference between classes exists across all layers, we first wish to perform a more thorough analysis of possible reasons for it.

\subsection{Semantics}
\label{ssec:comp-sem}

One possible explanation for the difference in BERT's treatment of blends and lexical compounds is that blends arise in lexical situations that are qualitatively different from those in which compounds are formed.
This would lead to a different distribution of semantic relationships between bases of blends and compounds.
In our annotated dataset we were able to witness such differences; for example, the \textsc{attribute} relation accounts for 23\% of compounds but 38\% of blends. 

\begin{figure}
    \centering
    \includegraphics[width=\linewidth]{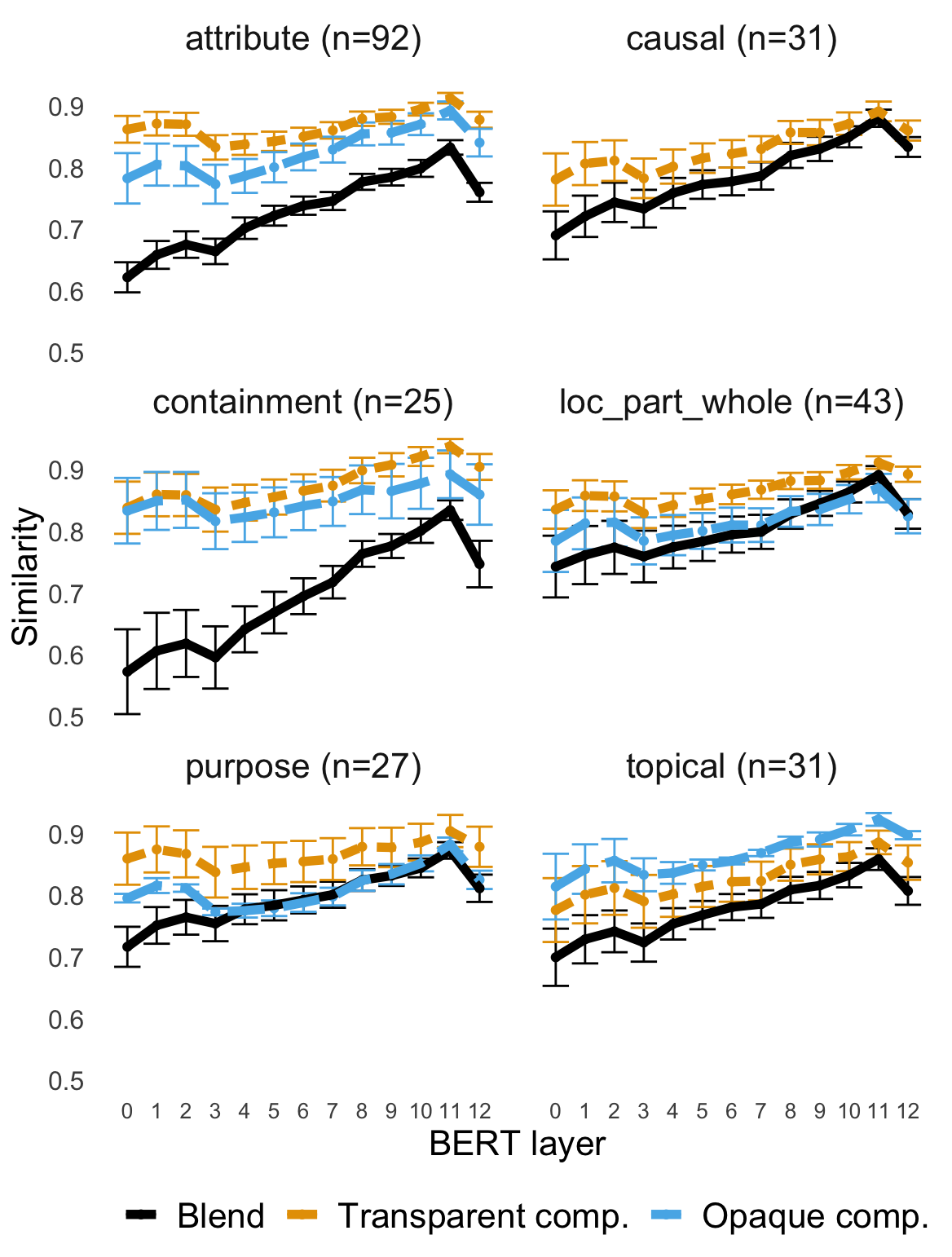}
    \caption{Pretrained BERT's similarity measures for each semantic relation with $n > 15$ instances.}
    \label{fig:per-cat-sims}
\end{figure}

If BERT's divergent treatment of blends and compounds is explained by the distribution over semantic categories for each complex word type, then we would expect the similarity scores \emph{within} categories to be identical.
Repeating the contextual similarity analysis within each semantic category, we find that there are substantial divergences between blends and compounds in several of the semantic categories.
\autoref{fig:per-cat-sims} presents the similarity scores for the six relations containing at least 15 observations; blend representations are less similar to their decomposed versions compared to compounds regardless of the relation.
A linear model trained to predict similarity
confirms that blends are less similar to their components than compounds ($\rho = -.128$, $p < .001$).

\subsection{Form}
\label{ssec:comp-smooth}
Another potential explanation is that differences in BERT's treatment of blends and lexical compounds are driven by the form of each compound, rather than the meaning.
On this view, the choice of whether to create a compound or a blend is a stylistic one \cite{LexicalBlendingasWordplay}, and so controlling for the character loss incurred in blends would produce the same processing difficulty for compounds.

\paragraph{Smoothies.}
If differences in surface form are what drives differences in contextualized representations, then transforming the compounds into mock-blends, which we term \say{smoothies}, should eliminate the differences between the two complex word types:
we would expect the function of the similarity of a contextual encoding of a blend to its bases given a context it naturally occurs in to be approximately the same function of similarity of a contextual encoding of a smoothie to its bases given the context the original compound occurs in.
We create our smoothies using \textsc{CopyCat}~\cite{kulkarni-wang-2018-simple}, a model which generates blends from two base forms via a sequence of character \texttt{copy} and \texttt{delete} actions learned over features extracted from an language model, an LSTM, and length-based heuristics.
We train an ensemble of 50 \textsc{CopyCat} models on the blends from~\newcite{deri-knight-2015-make} and apply them to our novel compounds.\footnote{We run the model ten times, and average the BERT differences over each base pair's resulting smoothies before aggregating for categories.}
Since \textsc{CopyCat} can produce only linear blends, we compare the BERT correspondence for smoothies against linear blends only (whose aggregate similarities are notably similar to those of blends as a whole).
In creating the smoothies,\footnote{Examples include \emph{bow} + \emph{person} = \emph{boerson} and \emph{junk} + \emph{time} = \emph{junime}.} we made sure that the overall rate of lost characters (\texttt{delete} operations) is comparable to that of the true linear blends.
We show in \autoref{fig:bert_sims}(b) that smoothies pose similar generalization challenges as blends: the gap between linear blends and smoothies is small, while generalization for smoothies is far below that of the original compounds. 

\paragraph{Tokenization.}
Having established that surface form is a main driver of the representational differences between blends and compounds, we now assess the specific impact of BERT's tokenization model, WordPiece (WP).
WordPiece is a trained model, consisting of a subword vocabulary constructed by identifying units (pieces) that appear repeatedly in a corpus.
It distinguishes between word-initial pieces, which may be whole words, and word-noninitial pieces which are marked by a special \say{\#\#} prefix.
A word is then assigned a sequence of pieces whose characters matches it when concatenated.
For example, WP(\say{segmenting})=[`segment', `\#\#ing'].
Such a model might be poorly suited to novel blends, which by definition reuse characters across bases, and which cannot be analyzed by traditional patterns of morphology.

To test the effect of segmentation, we provide WP with base-congruent segmentation points informed by their PAXOBS tags: for example, 
\textit{shoptics} is fed to BERT as $\rightarrow$ 
\textit{sh}+\textit{\#\#op}+\textit{\#\#tics}.
We find that this change does little to bridge the gap between blends and compounds: a redrawn version of \autoref{fig:bert_sims}(a) using this tokenization is almost identical to the original.
Upon further examination, we find that while pre-tokenizing with PAXOBS results in a larger number of wordpiece tokens (an average of $4.55$ vs. $3.30$), a similar leap occurs in compounds ($3.41$ vs. $2.48$), suggesting that WP does not produce morphologically accurate segments for compounds either~\cite{bostrom2020byte}. 
The crux of the issue must therefore lie within BERT.

\paragraph{}

In conclusion, we have shown that the root cause of blend mistreatment in large contextual transformer models is their form, although knowing only their sequence structure is not sufficient.
Therefore, in the following section we suggest models which attempt to identify blend segmentation points, but also ones which attempt recovery of their original bases, in order to place them in an appropriate topical context.

\section{Will it Unblend?}
\label{sec:unblend}

We next test to what extent existing models can help systems understand the meaning of novel blends, an aspect of human language understanding that has been little explored in NLP evaluation tasks.
As demonstrated in~\S\ref{sec:motiv}, successfully representing blends requires the capability to both properly decompose their form and identify the original constituents.
We therefore cast blend understanding as two tasks: segmenting blends into character sequences (\S\ref{ssec:segmentation}), and recovering blends' bases post-segmentation (\S\ref{ssec:recovery}).
We leave the task of recognizing blends to future work.\footnote{Note that the best OOV classification baseline in \citet{pinter2020nytwit}, a ridge classifier using character ngram features, reaches $.305$ F1 on the blends class (its macro-F1 on all 18 classes is $.323$).}

\paragraph{Compounds.}
Compounds were used as a comparative class in~\S\ref{sec:motiv}, but for the purpose of form understanding we focus on blends.
For compounds with a known segmentation, base recovery is trivial, as each side of the segmentation point is always a base.
As for segmentation, we have shown in~\S\ref{ssec:comp-smooth} that BERT's transformer layers are capable of recovering from poor WordPiece performance, and so the utility of segmenting compounds is limited compared to blends.\footnote{We nevertheless evaluated the segmenters on compounds (compare with \autoref{tab:seg_results}).
WP performs about as well as on blends in F1 ($.558$), and better in exact match ($34\%$), and Domain Unigram LM outperforms it on both ($.636$, $39\%$ respectively). In compounds, lenient and strict metrics converge.} 
Knowing that words are kept in their original form can define much simpler and more effective systems of discovery than the ones described below, such as a dictionary lookup of both sides for each possible single segmentation point.

\subsection{Blend Segmentation}
\label{ssec:segmentation}

We approximate the problem of inferring blend structure by defining a segmentation task over the character sequence which is the blend form, on the rationale that supplying a downstream system with character segments, each coherently representing a single known word or morpheme, would improve its ability to represent the input sequence.
For example, a character-aware system familiar with non-complex words might understand that the initial \emph{hat} from \emph{hatriotism} is related to \emph{hate} if given in isolation; but with \emph{hatr} it would be at a loss.

\paragraph{Metrics.} We draw on our PAXOBS schema (\S\ref{sec:data}) to define segment-level precision and recall scores for a given blend (e.g.~\emph{shoptics}: \texttt{AAXXBBBS}).
A system's prediction is a set of character indices where segmentation should occur.
We count any index which separates characters of the same label
as a false positive, towards precision (e.g.~the segmentation in [\textit{shopti};\textit{cs}]).
False negatives may be defined \textbf{strictly} or \textbf{leniently}: 
in the strict evaluation, a false negative is any segment that contains characters belonging to more than one base, or to a base and shared material (\texttt{X}), while a lenient evaluation permits the inclusion of shared material:
[\textit{shop};\textit{tics}] is leniently sound, but [\textit{sh};\textit{op};\textit{tics}] is strictly sound as well.
We report micro-level precision, as well as F1 computed with both lenient and strict recall, and lenient exact match.
We ignore prefixes and suffixes, and allow models to freely separate or include them in the adjacent base.

\paragraph{Systems.} We compare the following systems (see Appendix~\ref{app:seg} for implementation details):
\begin{itemize}[parsep=0pt,itemsep=2pt,leftmargin=2ex,labelwidth=2ex]
    \item \textbf{All-chars}. A baseline which marks every character as its own segment (perfect recall).
    \item \textbf{Sequence Tagger.} We annotated the 1,579 blends in \newcite{gangal2017charmanteau}'s dataset\footnote{After filtering duplicates from its 1,624 lines.}
    for PAXOBS tags and used them for training a supervised neural character-level tagger, whose results are converted into segmentations.
    \item \textbf{WordPiece.} We run WP \say{out of the box}.
    \item \textbf{In-domain Subwords.} We train BPE~\cite{sennrich-etal-2016-neural} and Unigram LM~\cite{kudo-2018-subword} subword tokenizers on news data from the Corpus of Contemporary American English~\cite[1990--2015;][]{davies-coca} using the sentencepiece package~\cite{kudo-richardson-2018-sentencepiece}, set to the same vocabulary size as the WP model.
    \end{itemize}
\begin{table}    
    \centering
    \small
    \begin{tabular}{lrcccr}
        \toprule
        Model & \#segs & Prec. & L F1 & S F1 & L EM \\
        \midrule
        All-chars & 10.15 & .272 & .427 & .427 & 0\% \\ 
        Seq. tagger & 4.70 & .291 & .400 & .376 & 5\% \\ 
        WordPiece & 3.30 & \textbf{.450} & \textbf{.562} & .484 & 22\% \\
        Domain BPE & 3.18 & .408 & .517 & .441 & \textbf{24}\% \\
        Domain ULM & 4.08 & .428 & .556 & \textbf{.492} & 22\% \\
        \bottomrule
    \end{tabular}
    \caption{Results for segmentation ($N=142$, micro-aggregation). \say{L} -- lenient, \say{S} -- strict, \say{EM} -- exact match. The desired number of segments varies between 2.05 and 3.40, depending on affixes and recall policy.}
    \label{tab:seg_results}
\end{table}

\paragraph{Results.}
The results in \autoref{tab:seg_results} show that all models struggle to find correct segmentation, even compared to the all-chars baseline.
The low performance of the supervised tagger suggests that little can be inferred from relative character placement, demonstrating the highly variable nature of novel blends.
Corpus-based segmentation models manage to segment over 20\% of the blends successfully.
This number is slightly higher when looking at the subset of linear blends: Domain BPE matches 29\% of them exactly.
Further analysis of the WP segmentations reveals a weakness in cases where the first post-\texttt{A} characters suggest a plausible continuation to base $A$, common enough to appear in WP's vocabulary, e.g.~[\emph{males};\emph{tream}] (true bases \emph{\textbf{male}}, \emph{main\textbf{stream}}; labels \texttt{XXAABBBBBB}), or [\emph{chip};\emph{ster}] (true bases \emph{\textbf{chi}cano}, \emph{h\textbf{ipster}}; labels \texttt{AXXBBBBB}).

We next consider the challenge of reconstructing the base components for segmented blends.

\subsection{Blend Component Recovery}
\label{ssec:recovery}

We tasked different models with identifying the contributing bases ($A$, $B$)\footnote{In three-based blends, we denote the last base as $B$ for this task, surmising it is more important than the second base given the right-headedness tendency of English.} out of all possible words given a gold-segmented blend and an input vocabulary.
We create sets of candidate words for each blend which could, in principle, create the same blend as the true bases.
For example, the blend \textit{thrupple} = \textit{\textbf{thr}ee + cou\textbf{ple}} will induce candidates such as \textit{\textbf{thr}ash} for $A$ and \textit{exam\textbf{ple}} for $B$.
We report the following metrics:
\begin{itemize}
    \item \textbf{MRR-$(A,B,\omega)$} is the mean reciprocal rank of true base $A$/$B$ across all possible candidates for that side (\textit{single-side prediction}), or ($\omega$) of the true base pair out of all possible base pair candidates (\textit{pair prediction});
    \item \textbf{Precision @1} is the proportion of blends for which the top candidate is the true base pair.\footnote{This contrasts with precision as reported by \newcite{cook-stevenson-2010-automatically}, who count a pair as correct if at least \textit{one} base is correct.}
\end{itemize}

In order to maintain a fair comparison between the models (see below), we extracted the candidate lists for all model evaluations from the GloVe~\cite{pennington-etal-2014-glove} model's vocabulary, as it is the only one restricted for in-vocabulary testing.
In total, 33 of the candidate lists (12\%) are singletons, including two blends where neither base has negative samples.
Six blends (4\%) lacked the correct base for one of the sides, and these cases were treated as ranked last among candidates; three of these lists were empty, translating to in a \#1 rank for all systems.
The lower bounds on the metrics resulting from these candidate list limitations are presented at the top of \autoref{tab:recovery_results}.\footnote{The full lists are available on the repository.}

\paragraph{\textsc{BERT ranker}.}
We propose a contextual representational approach for ranking two-sided base candidates using iterative piece prediction: we replace each appearance of a blend $b$ in its context sentence $(w_1, \dots, w_{i-1}, b, w_{i+1}, \dots, w_n)$ with two successive \texttt{[MASK]} tokens: $(w_1, \dots, w_{i-1}, m_1, m_2, w_{i+1}, \dots, w_n)$.
Then, we use a pretrained BERT masked language model to compute wordpiece prediction distributions for these masked tokens.
We sort all possible candidate base pairs $\langle l,r \rangle$ according to the sum of probabilities for their bases' first pieces, $P(m_1=l_0)+P(m_2=r_0)$,\footnote{This is crucial, since predicting a `\#\#'-initial suffix token effectively attaches it to the preceding token} and record the rank of the true base pair.
When candidate pairs have the same first-piece pair, we break ties by iteratively predicting the next pieces after inserting the shared pieces into the input.
We also implement an ablation (\textsc{$-$context}) where no context is added to the masks, in order to evaluate the contribution of the sentence contexts in isolation.

For single-side metrics, we report the rank of the true base in the prediction distribution of a single \texttt{[MASK]} token (instead of two); in another variant (\textsc{$+$other-base}) we add the true base from the other side to the context, in order to level the playing field with the baselines, which we describe next (see Appendix~\ref{app:rec} for implementation details):
\begin{itemize}[parsep=0pt,itemsep=2pt,leftmargin=2ex,labelwidth=2ex]
    \item \textbf{Character RNN}.
    We separately train a forward and a backward character-level RNN on over 100,000 documents from the Westbury corpus \cite{shaoul2010westbury}.
    We feed the blend's left (right) context to the forward (backward) RNN, then record the probability of each $A$ ($B$) candidate as a continuation of the context, computed as the average of character log-likelihoods.
    \item \textbf{Edit distance (ED)}.
    Following \newcite{cook-stevenson-2010-automatically}, we compare the string similarity (Levenshtein distance) between base candidate pairs' orthographic forms.\footnote{A variant using phonological forms, extracted from a phonological lexicon~\cite{lee-etal-2020-massively}, was limited by only having pronunciations for a fraction of bases and candidates. In cases where both base pronunciations were found the ranking was good, hinting at a promising avenue for future work by implementing automatic text-to-phone modeling.}
    Single-side prediction fixes one base and ranks candidates from the other side based on similarity.
    \item \textbf{Static embeddings}.
    We calculate cosine similarity between candidate base pairs' embeddings in fastText~\cite{mikolov2018advances} and GloVe~\cite{pennington-etal-2014-glove}.
    fastText includes character n-grams, allowing an assessment of the utility of subword information.
\end{itemize}

To summarize, both \textbf{ED} and \textbf{Static} methods are contextless pair-matchers which operate in the \textsc{$+$other-base} knowledge setup when evaluated for MRR-$A$ and MRR-$B$; \textbf{Character RNN} is a single-base ranker which uses context from one side only and cannot be helped by knowledge of the other base.

\begin{table}
    \small
    \centering
    \begin{tabular}{lHHcccc}
        \toprule
         &  &  & \multicolumn{3}{c}{MRR-} & \\
        \cmidrule(lr){4-6}
        Model & $N$ & Vocab & $A$ & $B$ & $\omega$ & P@1\\
        \midrule
        Lower bound & 142 & 840B & .115\phantom{$^*$} & .257\phantom{$^*$} & .036 & .014 \\
        \midrule
        Character RNN & 142 & 840B & .162\phantom{$^*$} & .368\phantom{$^*$} & .060 & .021 \\ 
        Edit distance & 142 & 840B & .176$^*$ & .432$^*$ & .066 & .014 \\
        fastText & 142 & 840B & .357$^*$ & .610$^*$ & .167 & .127 \\
        GloVe & 142 & 840B & .449$^*$ & .734$^*$ & .188 & .127 \\ 
        \midrule
        \textsc{BERT ranker} & 142 & 840B &  .392\phantom{$^*$} & .711\phantom{$^*$} & \textbf{.288} & \textbf{.264} \\
        \phantom{BE}\textsc{$+$other-base} & 140 & 840B &  .403$^*$ &  .703$^*$ \\
        \phantom{BE}\textsc{$-$context} & 142 & 840B &  .379\phantom{$^*$} &  .675\phantom{$^*$} & .147 & .127 \\ 
        \phantom{BERT}\textsc{$+$other-base} & 142 & 840B &  .379$^*$ &  .668$^*$ \\
        \bottomrule
    \end{tabular}
    \caption{Results for component recovery.
    $^*$Results dependent on knowledge of the correct base on the other side.}
    \label{tab:recovery_results}
\end{table}

\paragraph{Results.}
Results are presented in \autoref{tab:recovery_results}.
We note the higher performance on $B$ bases achieved by all models, a fact which advantages WordPiece which leaves word-initial pieces unmarked (see~\S\ref{ssec:comp-smooth}), as opposed to models such as XLM~\cite{conneau2019unsupervised}, which mark word-final pieces.
If the beginning of the blend bears more resemblance to the base it originated from, there's a better chance of properly representing that base in the overall blend.
These findings suggest an iterative setup, where first the $B$ base is predicted and only then $A$ is matched, might prove more successful.
We leave this variant to future work.

Our \textsc{BERT ranker} model outperforms all baselines in the more realistic full-word setting (MRR-$\omega$, P@1).
When ranking single bases, it does not benefit much from awareness of the true other base (the oscillations recorded in the table are too small to be meaningful), suggesting that most of its power lies in processing context and not in word form representation.
This conclusion is further supported by the superior performance of the static type-level GloVe embeddings, whose lead over fastText and \textsc{BERT$-$context} in all MRR measures suggests that word form is less helpful even in uncontextualized settings.
The particularly poor performance of the character RNN and edit distance model shows that it is difficult to learn the task without any semantic signal.

\paragraph{Error Analysis.}
A qualitative assessment of the contexts which help \textsc{BERT ranker} to predict bases perfectly relative to the \textsc{$-$context} variant shows that they typically contain one or more of the bases in their entirety (e.g.,~\emph{eggcessories} appears near multiple occurrences of the word \emph{eggs}).
By contrast, in some longer contexts containing diverse topics, the inclusion of context wipes out the accessibility of the component bases, typically the first one (e.g.~\emph{chesticle}, in which the context does not mention body parts, or \emph{cancerchondria} which mentions the word \emph{condition} but neither of the bases).

\section{Related Work}

Prior work on blends has largely focused on \textbf{generation}~\cite[e.g.,][]{das2017neuramanteau,simonentendrepreneur,deri-knight-2015-make,kulkarni-wang-2018-simple,smith2014nehovah}.
While \newcite{gangal2017charmanteau} provide a unified dataset of 1,579 blends, annotated for bases, they do not provide contexts for real-world appearances of the blends, nor a breakdown of the semantic relationship between their constituents.
Moreover, some are synthetically generated by a seq2seq model.
In addition, these works all restrict their models to linear two-word blends.
Our PAXOBS scheme handles nonlinear and multi-base blends.

\newcite{cook-stevenson-2010-automatically} presented a noncontextual method for blend base detection using a dictionary-based lexicon, evaluated over an unreleased dataset, and \newcite{ek2018identifying} used features from static embeddings to unblend words in Swedish.
We adopt the candidate-ranking approach of these works to evaluate component recovery, but incorporate context with context-sensitive language models, and add the task of blend segmentation.

Extracting the semantics of constituents from larger phrases is not a problem unique to single-token blends.
\newcite{shwartz-waterson-2018-olive} worked on multi-word compounds; \newcite{maddela-etal-2019-multi} segment hashtags, roughly half of which are akin to our notion of compounds, by training a neural scoring system over features extracted from word form, dictionary lookup and language model probabilities.
Another connection is to the learning of morphological rules, e.g., for processes such as derivation~\cite{kondratyuk-2019-cross} and lemmatization~\cite{chrupala2006simple, Ullman:1976:BCL:321921.321922, Hirschberg:1977:ALC:322033.322044}.
\newcite{cotterell-schutze-2018-joint} present a supervised model of derivational morphology that jointly accounts for segmentation as well as composition of static word embeddings from the embeddings of morphemes, thereby touching on two of the main tasks undertaken in our paper. However, the application of such a model to blends is complicated by the relative lack of labeled training data, as well as the irregularity of the underlying phenomenon.

Novel blends are an example of linguistic creativity, which frequently operates at the subword level.
Related phenomena include \textbf{eggcorns}, which are alternative spellings that yield an apparently more transparent relationship between form and function~\cite{reddy2009understanding}; \textbf{puns}, which substitute words in new contexts based on phonological similarity~\cite{jaech2016phonological}; respellings that attempt to reintroduce prosodic expression into spelling~\cite{brody2011cooooooooooooooollllllllllllll}; intentional obfuscation \cite{zalmout-etal-2019-unsupervised}; and typographical errors~\cite{heigold2017robust}. We therefore view blends as an instance of a broad set of creative phenomena that poses challenges for the token-based approaches that currently dominate natural language processing.

\section{Conclusion}
\label{sec:concl}

This work focuses on the challenge of interpreting novel blends, which requires integrating subword structure and contextual features.
We present a new dataset annotated using a novel character-level schema as well as for semantic tags, and offer preliminary evaluations showing that (a) blends are handled differently than compounds by BERT, due mostly to the phenomenon of character loss; (b) existing tokenizers generally do not respect blend boundaries; and (c) recovering the components of a blend is a difficult task which challenges word-form, distributional, and contextual approaches.
Our results further highlight that annotation schemata such as those of \newcite{tratz-hovy-2010-taxonomy}, which were designed for noun compounds, are generalizable to other relational word types.
In future work, we plan to integrate these signals into a better blend processor, and to further address the effect of blends on downstream tasks from semantic and syntactic viewpoints.
In addition, we aim to further examine the methods from our experiments on other classes of novel words and in other languages.
We also plan to add phonetic resources for improving treatment of nonlinear blends.

\ifaclfinal
    \section*{Acknowledgments}
    We thank Max Bittker for providing us with details and logs from the NYT bot.
    We thank Lelia Glass, Kyle Gorman, Arya McCarthy, Ian Stewart, Kristina Toutanova, Sarah Wiegreffe, Yuwei Wu, Wei Xu, Diyi Yang, and the anonymous reviewers for their valuable notes.  
    YP is a Bloomberg Data Science PhD Fellow. 
    CJ is a postdoctoral research associate supported by NSF grant 1849236 awarded to Maryellen MacDonald.
\fi

\bibliography{unblend,anthology}
\bibliographystyle{acl_natbib}

\clearpage

\appendix

\section{\say{Will it Unblend?}: Supplementary}
\label{sec:appendix}

\subsection{Annotation}
\label{app:ann}

When annotating the PAXOBS schema, the \texttt{A} base is defined as the one whose exclusive material precedes all other exclusive materials in the blend, and from then on iteratively through the alphabet.

\subsection{Segmentation Experiment}
\label{app:seg}

\paragraph{Character Tagger.}
We manually annotated 550 of the 1,579 blends in the~\cite{gangal2017charmanteau} dataset, and passed the rest through a heuristic program to cover whatever linear blends remained.
The program flagged 150 blends as suspected nonlinear, so we manually annotated them as well.
A tagger trained on only the 550-blend set originally annotated did not reach better F1 scores than the one reported in the main text, trained on the full set.

We use a 3-layer Bidirectional LSTM followed by a 4-layer MLP (tuned in the 2--4 range on dev) with ReLU activation (tuned vs. tanh), trained for 30 epochs with early stopping, optimized using Adam~\cite{kingma2014adam} with learning rate 0.01 and a batch size of 96 (not tuned).
The character embedding dimension is 200 and the hidden dimensions for the LSTM and MLP are both 192 (not tuned).
Tagging accuracy on the dev set is $.457$ and on our dataset $.462$.
We translate the resulting PAXOBS tags into segmentations by segmenting on each label change (so $\langle$\say{shoptics}, \texttt{AAXXBBBS}$\rangle$ becomes \say{\textit{sh};\textit{op};\textit{tic};\textit{s}}).
The model is implemented in PyTorch~\cite{pytorch}.

\paragraph{Domain models.}
The vocabulary size of BPE and the Unigram LM are by default automatically inferred, but may be specified.
For BPE, we tested several vocabulary sizes: 20,000, 30,000, 30,522 (WP vocabulary size) and 40,000, selected heuristically, and 30,522 performed best.
Consequently, we used the same size for Unigram LM.

\paragraph{Results.}
Strict exact match scores are: All-chars, 0.0\%; Seq. tagger, 3.5\%; WordPiece, 7.7\%; Domain BPE, 10.6\%; Domain ULM, 6.3\%.  This ordering is consistent with the lenient exact match scores reported in \autoref{tab:seg_results}.

\subsection{Recovery Experiment Details}
\label{app:rec}

A single hyperparameter controlling the minimum length of candidate overlap in linear blends was set to 3 with no tuning.
Candidates were selected according to bases' stemmed form.
We provide the complete candidate lists in the project repository.

\paragraph{Character RNN.} We use 2-layer GRUs with embedding dimension 128 and hidden dimension 256 (all chosen manually with no tuning), and a sample of $\sim$109,000 documents from the Westbury corpus \cite{shaoul2010westbury}, and optimize using Adam with early stopping determined by performance on a held-out development set of 10,000 randomly sampled documents.
The model is implemented in PyTorch.

\paragraph{FastText.} We used the Engish CommonCrawl 300-dimension vectors available from \url{https://fasttext.cc}, inferring OOV words using FastText software.

\paragraph{GloVe.} We used 300-dimensional GloVe vectors trained on the 840-billion CommonCrawl corpus, obtained from \url{https://nlp.stanford.edu/projects/glove/}.

\paragraph{BERT.} In candidate pair ranking, when multiple candidate pairs have the same initial wordpieces $\langle$ $l_0$=\textit{pre}, $r_0$=\textit{suf} $\rangle$, we create a new sentence input \say{\textit{left\_context} \textit{pre} [MASK] \textit{suf} [MASK] \textit{right\_context}} and continue predicting the following piece pair, $\langle l_1, r_1 \rangle$ iterating until there are no more ties.
At any point in the process, candidates which run out of wordpieces are floated to the top of the working ranked list by base order ($A$-ending before $B$-ending).

We inferred the words in context based on the \textsc{bert-base-uncased} \textsc{BertForMaskedLM} module obtained via  \url{https://github.com/huggingface/transformers} (version 2.0.0).
This required lowercasing all input prior to processing.

\end{document}